\documentclass[journal]{IEEEtran}
\usepackage{multirow}
\usepackage{cite}
\usepackage{amsmath,amssymb,amsfonts}
\usepackage{algorithmic}
\usepackage{textcomp}
\usepackage{array}
\usepackage{booktabs}
\usepackage{color}
\usepackage[numbers,sort&compress]{natbib}

\usepackage{subfigure}
\usepackage{caption2}
\usepackage[pdftex]{graphicx}

\usepackage{stfloats}
\begin{document}
\title{Bilinear Supervised Hashing Based on 2D Image Features}
\author{Yujuan~Ding, Wai Kueng Wong, Zhihui~Lai and Zheng~Zhang 
\thanks{Y. Ding and W. K. Wong are with the Institute of Textiles and Clothing, The Hong Kong Polytechnic University, Hong Kong (e-mail: dingyujuan385@gmail.com; calvin.wong@polyu.edu,hk).}
\thanks{Z. Lai is with the College of Computer Science and Software Engineering, Shenzhen University, Shenzhen 518060, China, and also with the Institute of Textiles and Clothing, The Hong Kong Polytechnic University, Hong Kong (e-mail: lai\_zhi\_hui@163.com).} 
\thanks{Zheng Zhang is with School of Information Technology \& Electrical Engineering, The University of Queensland, Brisbane, QLD 4072, Australia. (e-mail:darrenzz219@gmail.com).}}


\maketitle

\begin{abstract}
Hashing has been recognized as an efficient representation learning method to effectively handle big data due to its low computational complexity and memory cost. Most of the existing hashing methods focus on learning the low-dimensional vectorized binary features based on the high-dimensional raw vectorized features. However, studies on how to obtain preferable binary codes from the original 2D image features for retrieval is very limited. This paper proposes a bilinear supervised discrete hashing (BSDH) method based on 2D image features which utilizes bilinear projections to binarize the image matrix features such that the intrinsic characteristics in the 2D image space are preserved in the learned binary codes. Meanwhile, the bilinear projection approximation and vectorization binary codes regression are seamlessly integrated together to formulate the final robust learning framework. Furthermore, a discrete optimization strategy is developed to alternatively update each variable for obtaining the high-quality binary codes. In addition, two 2D image features, traditional SURF-based FVLAD feature and CNN-based AlexConv5 feature are designed for further improving the performance of the proposed BSDH method. Results of extensive experiments conducted on four benchmark datasets show that the proposed BSDH method almost outperforms all competing hashing methods with different input features by different evaluation protocols.  
\end{abstract}


\IEEEpeerreviewmaketitle

\section{Introduction}
\IEEEPARstart{H}{ashing} has been a hot research topic on image processing and pattern recognition in the past decade due to its promising performance in image and video storage and retrieval \cite{wang2018,chen2018,zhang2017}. Hashing aims to find a series of hash functions to map the high-dimensional image features into a low-dimensional hamming space. The final purpose of hashing is to learn good binary codes that can preserve the similarity correlations and well represent images with the minimum information loss.

Learning to hash method is also known as data-dependent hash codes learning. It has become a substantial part of hashing research and attracted much attention owing to its promising performance \cite{zhang2018,Rshen2017,shen2018,shen2016}. Its main objective is how to design the hashing function to minimize the distance (loss) between the low-dimensional hashing features (binary codes) and the original high-dimensional features, or how to preserve similarities between them. To this end, a number of distance and similarity measurements were proposed, including minimizing the product of the similarity and distance, maximizing the similarity-similarity product, minimizing the distance-distance difference, and etc. \cite{kulis_learning,shen_learning,zhang_marginal,zhang_discriminative}. Quantization error minimization is a widely-adopted strategy for objective function design as a special form of distance-distance difference minimization. Existing studies claim that using quantization method can obtain high-quality hash codes that can achieve better performance for image retrieval task.
\par Although a large number of hashing methods have been proposed with different objective functions, most of them are still based on linear projection and only focus on high-dimensional vectorized features \cite{zhao2014}. Even though some 2D features are adopted, they are still manually converted into vectors in practice. However, such operation fail to capture the structural information contained in matrix-form features. Actually, natural matrix structures are frequently observed in image features such as vector of locally aggregated descriptors (VLAD), Fisher Vector and Histogram of oriented gradient (HOG) \cite{jegou,nchez2013,dalalcvpr}. These features carry abundant information in matrix form. As introduced in Kim's work \cite{kim_blsh}, the residual between matrix features and their nearest visual vocabularies is encoded into a matrix form. As a result, manually vectorizing these matrices may lose structural information and weaken their performance. Moreover, when 2D matrix feature is re-organized as a high-dimensional vector feature, the high storage and time complexity cause a big problem for large projection construction. Therefore, a better choice is to develop 2D hashing function learning to take advantage of small size matrix-form features, resulting in better performance.
\par Bilinear models have been successfully used in other applications such as 2D-PCA, 2D-CCA, 2D-LDA \cite{yang2004,lee2007,ye_two} to directly process 2D inputs, the results of which showed that better performance can be achieved via bilinear learning. Besides, the well-known convolutional neural network (CNN) has shown exceptional performance on image analysis tasks. The advancement, to most degree, benefits from the convolutional layers where the convolution operation can capture local dependencies in the original image and preserve the spatial relationship between pixels. The success of CNN and its application in different areas shows the importance of containing spatial information in image analysis, and also indicates that it is promising in hashing to obtain high-quality binary features if a suitable objective function can be designed to take the advantages of the matrix form features. 
\par Recently, a few works have paid attention to 2D image features and some bilinear hash learning methods were proposed \cite{kim_blsh,liu2016,gong_bpbc}. One representative work \cite{gong_bpbc} is the bilinear projection based binary codes (BPBC) learning method, which formulates the objective function by minimizing the angle between the rotated feature vector and its binary codes. Specifically, two rotation matrices are employed to rotate descriptor matrix for generating effective binary codes. However, since BPBC is an unsupervised method, the label information is not considered, yielding inferior retrieval performance. Moreover, since its main idea is to strengthen the original feature using rotation strategy, the performance is degraded when the binary codes are short. Similarly, other methods \cite{kim_blsh,liu2016} also did not achieve very satisfactory results either. 
\par Due to the inspiration and limitation of the current research mentioned above, in this paper, we propose a bilinear supervised discrete hashing method (BSDH) based on quantization loss minimizing to learn binary codes from 2D image features. This method utilizes bilinear projection to binarize matrix features which can keep the image matrix’s intrinsic space relationship and integrates the idea of bilinear projection approximation and vectorization binary codes regression. The main contributions of this paper are summarized as follows:
\par 1)	We propose a 2D feature-based hashing method in which the design of the objective function is based on matrices instead of vectors. This design can effectively avoid the loss of crucial spatial information of images contained in 2D features. The proposed method integrates the bilinear projection approximation and supervised linear regression, so that high-quality binary codes are promised to be learned. 
\par 2)	A novel optimization strategy based on bilinear discriminant analysis and discrete optimization is developed. The discrete optimization strategy is an alternatively iterative method, which minimizes the quantization loss of the learned binary codes and the projected real value matrix in the regression to generate high-quality binary codes. 
\par 3)	We explore two kinds of feature extraction methods to extract 2D image features for the proposed hashing method, i.e., non-deep FVLAD (Fast VLAD) feature and deep AlexConv5 feature. The FVLAD feature is designed based on SURF (speed up robust feature) feature as an efficient alternative to VLAD, while the AlexConv5 feature is extracted from the last convolutional layer of the typical CNN, i.e., AlexNet. According to experimental results, the proposed BSDH obtains higher recall rates and MAPs against other methods. Further, with AlexConv5, the performance of BSDH is significantly enhanced.  
\par The rest of this paper is organized as follows. In section  \uppercase\expandafter{\romannumeral2}, related studies about supervised discrete hashing and bilinear projection binary codes learning are briefly reviewed. In section \uppercase\expandafter{\romannumeral3}, our bilinear supervised discrete hash learning model is proposed. Section \uppercase\expandafter{\romannumeral4} introduces the discrete optimization algorithm we adopted to solve the optimization model. Experiments are explained in section \uppercase\expandafter{\romannumeral5} to evaluate the performance of BSDH. Finally, conclusions are given in section \uppercase\expandafter{\romannumeral6}.

\section{related work}
In this section, we briefly review some related studies about bilinear projection based binary codes and supervised discrete hashing. 
\subsection{\textbf{Bilinear Projection Based Binary Codes}}
The bilinear projection based binary codes (BPBC) method is a bilinear binary code learning method which explores the feature enhancement for better retrieval and classification performance using the matrix rotation strategy. Specifically, the high-quality binary codes are obtained by bilinear rotating the original feature matrix without dimensionality reduction with two random orthogonal matrices as follows:
   \begin{equation}
   \label{eq_1}
   X\to R_{1}^{T}X{{R}_{2}},
   \end{equation}
where $X$ is the original data and ${{R}_{1}}$, ${{R}_{2}}$ are two rotation matrices. Combined with threshold and vectorization, the final bilinear code is generated by: 
\begin{equation}
\label{eq_2}
    H\left( X \right)=\operatorname{vec}\left( \operatorname{sgn} \left( R_{1}^{T}X{{R}_{2}} \right) \right),
\end{equation}                     
where $ \operatorname{sgn} \left( \centerdot  \right)$ is the sign function and $\operatorname{vec}\left( \centerdot  \right)$ denotes vectorization. To learn the binary codes, the authors chose to minimize the angle ($\theta $) between the rotated feature vector $\operatorname{vec}\left( R_{1}^{T}X{{R}_{2}} \right)$ and its binary encoding $\operatorname{vec}\left( \operatorname{sgn} \left( R_{1}^{T}X{{R}_{2}} \right) \right)$:

\begin{align}\small
\label{eq_3}
  & \sum\limits_{i=1}^{n}{\cos ({{\theta }_{i}})} =\sum\limits_{i=1}^{n}{\left( \frac{\operatorname{vec}{{\left( \operatorname{sgn} \left( R_{1}^{T}{{X}_{i}}{{R}_{2}} \right) \right)}^{T}}}{\sqrt{d}}\operatorname{vec}\left( R_{1}^{T}{{X}_{i}}{{R}_{2}} \right) \right)} \nonumber\\ 
 & =\frac{1}{\sqrt{d}}\sum\limits_{1}^{n}{\operatorname{tr}\left( {{B}_{i}}R_{2}^{T}X_{i}^{T}{{R}_{1}} \right)}.
\end{align}

The final objective function turns into:
\begin{equation}
\begin{aligned}
\label{eq_4}
  & Q\left( B,{{R}_{1}},{{R}_{2}} \right)=\underset{B,{{R}_{1}},{{R}_{2}}}{\mathop{\max }}\,\sum\limits_{i=1}^{n}{\operatorname{tr}\left( {{B}_{i}}R_{2}^{T}X_{i}^{T}{{R}_{1}} \right)} \\ 
 & \begin{matrix}
   s.t. & {{B}_{i}}  \\
\end{matrix}\in {{\left\{ -1,+1 \right\}}^{{{d}_{1}}\times {{d}_{2}}}},R_{1}^{T}{{R}_{1}}=I,R_{2}^{T}{{R}_{2}}=I. \\ 
\end{aligned}
\end{equation}

The objective function can be easily solved by alternatively updating variables. Specifically, ${{R}_{1}}$ and ${{R}_{2}}$ are obtained by singular value decomposition (SVD), and low dimensional binary codes can be learned if ${{R}_{1}}$ and ${{R}_{2}}$ are downsized. However, the bilinear rotation discussed in this research is without dimension reduction. Even though authors considered learning low-dimensional binary codes, they only explored the performance of hash codes with half dimension as the original feature, while the learned binary codes are still highly dimensional. Based on our experimental results, its performance was degraded with the decrease of the binary code length, which is probably caused by not using any label information.

\subsection{\textbf{Supervised Discrete Hashing}}
Supervised discrete hashing (SDH) \cite{shen_sdh} attempts to use the manifold embedding strategy for formulating the hashing framework that can jointly learn a binary embedding and a linear classifier. Assuming that ${{x}_{i}}$ is the input descriptor vector, ${{y}_{i}}$ and ${{b}_{i}}$ are the label vector and binary codes respectively, $W$ denotes the mapping matrix for regression. The objective function of SDH is composed of ridge regression loss and embedding fitting error as follows:
\begin{equation}\small
\begin{aligned}
\label{eq_5}
  & \underset{B,W,F}{\mathop{\min}}\,\sum\limits_{i=1}^{n}{L\left( {{y}_{i}},{{W}^{T}}{{b}_{i}} \right)}+\lambda {{\left\| W \right\|}^{2}}+\upsilon \sum\limits_{i=1}^{n}{{{\left\| {{b}_{i}}-F\left( {{x}_{i}} \right) \right\|}^{2}}} \\ 
 & s.t.~~b_i = sgn \left( F\left(x_i \right) \right),i=1,...,n. 
\end{aligned}     
\end{equation}
The embedding function here is:
\begin{equation}
\label{eq_6}
    F\left( X \right)={{P}^{T}}\phi (X),
\end{equation}                                
where $\phi \left( x \right)$ is nonlinear anchor feature vector obtained by the RBF:
\begin{equation}\small
\label{eq_7}
\phi \left( X \right)={{\left[ \exp \left( {{\left\| X-{{a}_{1}} \right\|}^{2}}/\sigma  \right),\ldots ,\exp \left( {{\left\| X-{{a}_{m}} \right\|}^{2}}/\sigma  \right) \right]}^{T}},
\end{equation}
where ${{a}_{i}}$ and $\sigma $ are anchor points and kernel width. The kernel trick used here was widely used in kernel hash function e.g., BRE \cite{kulis_learning} and KSH \cite{liu_ksh,kulis_kernelized}. It has been proven that using kernel strategy is helpful to improve retrieval precision, but the computational cost will also increase. To solve the optimization problem in SDH, researchers kept the binary constraint and used discrete strategy to iteratively update variables. For the most challenging part in learning $B$, they kept the binary constraint and used the novel discrete cyclic coordinate descent method. 
\par This method formulated a joint objective function which leveraged label information and used discrete optimization method to solve the optimization problem. However, it only focused on vectorized features which cannot exploit the important structural information contained in 2D image features. Besides, the kernel embedding was time consuming to some extent, both effectiveness and efficiency were not promising.

\begin{figure*}[t]
\begin{center}
    \includegraphics[width=5.5in]{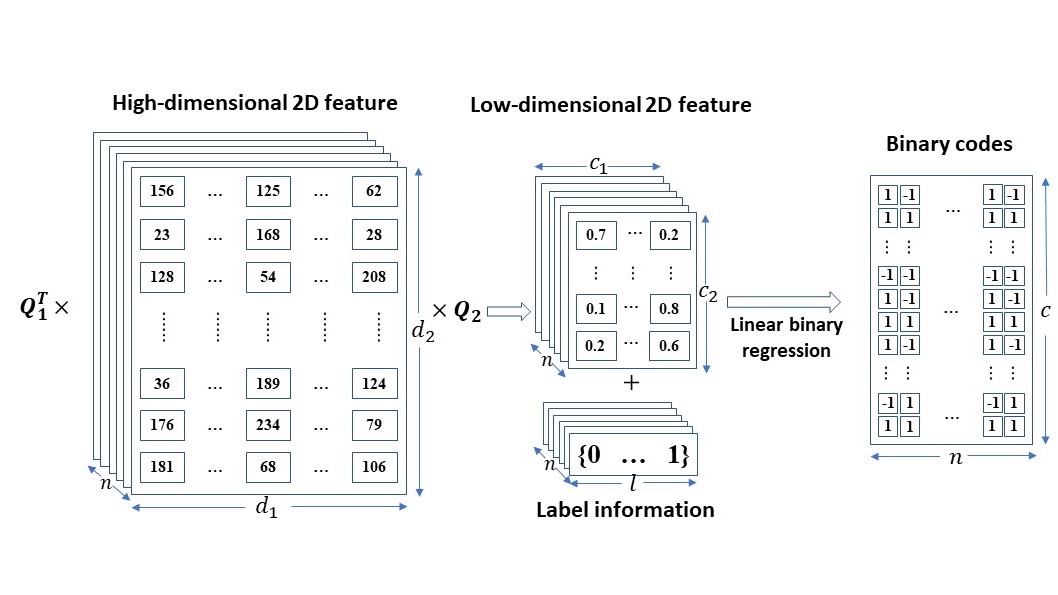}
    \caption{Brief pipeline of the proposed BSDH method.}
\end{center}
\label{fig1}
\end{figure*}

\section{Bilinear supervised discrete hashing method}
In this section, we firstly introduce how to employ the bilinear projection to obtain the low-dimensional representation of 2D image features. Then the supervised hashing stream is introduced to learn high-quality hash codes.

\subsection{\textbf{Motivation and General Idea}}
The hashing algorithm proposed in this paper is mainly inspired from two aspects. Firstly, as illustrated in section I, bilinear models show great potential to be used to develop well-performed hashing algorithm as they can directly deal with 2D features. It can effectively capture structure information contained in 2D image features and can avoid memory problem in calculation when applied on very big image datasets. However, existing bilinear hashing methods did not achieve remarkable performance since their objective functions were not well designed. For example, for BPBC, the objective function solely focused on minimizing the quantitation loss of the thresholding, which means that it only used bilinear mapping to make rotation of ordinary data for binary approximation, without exploiting any labels information and reconstruction information. Therefore, it is hard to obtain high-quality codes.
Secondly, as indicated in section II, the SDH achieved good performance in integrating  binary codes learning and image classification, which demonstrates that minimizing quantization loss and using discrete optimization strategy are effective. However, it is not efficient enough as the kernel embedding used for binary coding is time consuming and it becomes an obstacle for its application in big data scenarios. More importantly, the SDH method is based on vectorized features instead of 2D features. According to above discussion on the advantages of 2D features, applying 2D features is possible to further improve its performance. 
\par Based on above consideration, we propose a novel bilinear supervised discrete hashing (BSDH) method, which integrates the bilinear projection and vectorization binary codes regression. The idea of BSDH is to learn a bilinear mapping based on original image feature matrix so as to obtain low-dimensional 2D features and then to learn the binary codes by solving the supervised linear regression problem. Label information and reconstructive information are used in the proposed objective function simultaneously. Since solely ignoring the binary constraint or substituting it with relaxed function may degrade the performance, we adopted the discrete regularization learning in optimizing the proposed model. As such, the model can obtain binary codes with high performance by the 2D image features. 
\par Figure 1 illustrates the concept and principle of the proposed method. Firstly, we conduct bilinear projection on the raw image pixels or high-dimensional 2D features to obtain relatively low-dimensional 2D representation. Then we use the low-dimensional representation to further learn the compact binary code by supervised hash coding.

\subsection{\textbf{Low-dimensional 2D Feature by Bilinear Projection}}
For ease of presentation and understanding, notations used in the following section are listed in Table  \uppercase\expandafter{\romannumeral1}. 

\renewcommand\arraystretch{1.5}
\begin{table}[!t]
\hspace{0.5cm}
\setlength{\belowcaptionskip}{8pt}%
 \caption{\label{tab:test}Notations Introduction} 
 \begin{tabular}{p{40pt}p{40pt}p{120pt}}
  \toprule 
  Notation & Size & Description  \\ 
  \midrule 
  $ n $ &- &Number of samples\\
  $ c $ &- &Length of hash code\\
  $ l $ &- &Label dimension\\
  $ c_{1} $, $ c_{2} $ &- &Transition sizes for bilinear projection\\
  $ d_{1} $, $ d_{2} $ &- &Input sizes\\
 $ X_{i}$ &$ d_{1} \times d_{2}$ & Input feature matrix for ${{i}^{th}}$ image\\ 
 $ B=\{b_{i}\}_{n}$ & $c\times n$ & Hash code matrix\\
 ${{Q}_{1}}$, ${{Q}_{2}}$
& $d_{1}\times c_{1}$, $d_{2}\times c_{2}$
& Bilinear projection matrix used in BSDH\\
$Y={{\left\{ {{y}_{i}} \right\}}_{n}}$ & $l\times n$& Sparse label matrix for database\\
  \bottomrule 
 \end{tabular} 
\end{table}

Similar to BPBC, we use two matrices to conduct bilinear projection on high-dimensional 2D feature ${{X}_{i}}$ as ${{Q}_{1}}^{T}{{X}_{i}}{{Q}_{2}}$, but the difference is that the matrices are in smaller sizes. With ${{Q}_{1}}\in {{R}^{{{d}_{1}}\times {{c}_{1}}}}$ and ${{Q}_{2}}\in {{R}^{{{d}_{2}}\times {{c}_{2}}}}$, the input feature can be downsized from ${{d}_{1}}\times {{d}_{2}}$ to ${{c}_{1}}\times {{c}_{2}}$. Here we name ${{c}_{1}}$ and ${{c}_{2}}$ transition sizes. Inspired by linear discriminant analysis (LDA) [19], we consider the inter-class scatter matrix ${{D}_{b}}$ and intra-class scatter matrix ${{D}_{w}}$ as follows:
\begin{equation}\small
\begin{aligned}
\label{eq_8}
{{D}_{b}} &=\sum\limits_{i=1}^{l}{{{a}_{i}}{{\left\| {{Q}_{1}}^{T}\left( {{M}_{i}}-{{M}_{0}} \right){{Q}_{2}} \right\|}^{2}}}\\
&=tr\left( \sum\limits_{i=1}^{l}{{{a}_{i}}{{Q}_{1}}^{T}\left( {{M}_{i}}-{{M}_{0}} \right){{Q}_{2}}Q_{2}^{T}{{\left( {{M}_{i}}-{{M}_{0}} \right)}^{T}}{{Q}_{1}}} \right),
\end{aligned}
\end{equation}
\begin{equation}\small
\begin{aligned}
\label{eq_9}
&{{D}_{w}}=\sum\limits_{i=1}^{l}{\sum\limits_{{{X}_{j}}\in {{\Pi }_{i}}}{{{\left\| {{Q}_{1}}^{T}\left( {{X}_{j}}-{{M}_{i}} \right){{Q}_{2}} \right\|}^{2}}}}\\
&=tr\left( \sum\limits_{i=1}^{l}{\sum\limits_{{{X}_{j}}\in {{\Pi }_{i}}}{{{Q}_{1}}^{T}\left( {{X}_{j}}-{{M}_{i}} \right){{Q}_{2}}{{Q}_{2}}^{T}{{\left( {{X}_{j}}-{{M}_{i}} \right)}^{T}}{{Q}_{1}}}} \right),
\end{aligned}
\end{equation}
where ${{a}_{i}}$ is the number of samples of ${{i}^{\text{th}}}$ class, ${{M}_{i}}$ and ${{M}_{0}}$ are the mean of input for ${{i}^{\text{th}}}$ class and the mean of input for the whole dataset. By simple formulation, ${{D}_{b}}$ and ${{D}_{w}}$ can be further transformed into:
\begin{equation}
\label{eq_10}
{{D}_{b}}=tr\left( {{Q}_{1}}^{T}S_{b}^{{{Q}_{2}}}{{Q}_{1}} \right),
\end{equation}
\begin{equation}
\label{eq_11}
{{D}_{w}}=tr\left( {{Q}_{1}}^{T}S_{w}^{{{Q}_{2}}}{{Q}_{1}} \right),
\end{equation}
where $S_{b}^{{{Q}_{2}}}$, $S_{w}^{{{Q}_{2}}}$ are defined as follows:
\begin{equation}
\label{eq_12}
S_{b}^{{{Q}_{2}}}=\sum\limits_{i=1}^{l}{{{a}_{i}}\left( {{M}_{i}}-{{M}_{0}} \right){{Q}_{2}}{{Q}_{2}}{{\left( {{M}_{i}}-{{M}_{0}} \right)}^{T}}},
\end{equation}
\begin{equation}
\label{eq_13}
S_{w}^{{{Q}_{2}}}=\sum\limits_{i=1}^{l}{\sum\limits_{{{X}_{j}}\in {{\Pi }_{i}}}{\left( {{X}_{j}}-{{M}_{i}} \right){{Q}_{2}}{{Q}_{2}}^{T}{{\left( {{X}_{j}}-{{M}_{i}} \right)}^{T}}}}.
\end{equation}
To make the bilinear projection with minimum information loss, we aim to maximize the between-class distance and minimize within-class distance, which equals to maximize the following function:
\begin{equation}
\label{eq_14}
\begin{aligned}
&\max \left( {{\left( D_{w}^{{}} \right)}^{-1}}D_{b}^{{}} \right)\\
=&\max \left( tr{{\left( {{Q}_{1}}^{T}S_{w}^{{{Q}_{2}}}{{Q}_{1}} \right)}^{-1}}tr\left( {{Q}_{1}}^{T}S_{b}^{{{Q}_{2}}}{{Q}_{1}} \right) \right),
\end{aligned}
\end{equation}
where ${{Q}_{1}}$ can be obtained by executing eigen decomposition on the matrix ${{\left( S_{w}^{{{Q}_{2}}} \right)}^{-1}}S_{b}^{{{Q}_{2}}}$.
Similarly, we can obtain ${{Q}_{2}}$ by solving the eigen decomposition problem of ${{\left( S_{w}^{{{Q}_{1}}} \right)}^{-1}}S_{b}^{{{Q}_{1}}}$. After obtaining ${{Q}_{1}}$  and ${{Q}_{2}}$ , we can calculate the bilinear-projected matrix $H$ for further updating of other variables by the following equations.
\begin{equation}
\label{eq_15}
{{h}_{i}}=\operatorname{vec}\left( Q_{1}^{T}{{X}_{i}}{{Q}_{2}} \right),
\end{equation}

\begin{equation}
\label{eq_16}
H=[{{h}_{1}},{{h}_{2}},\ldots ,{{h}_{n}}]. 
\end{equation}            

\subsection{\textbf{Bilinear Supervised Discrete Hashing}}
To achieve better binary fitting based on the obtained bilinear feature matrix $H$, we further impose the linear projection on the matrix and minimize the quantity loss for the binary codes as follows:
\begin{equation}
\label{eq_17}
\begin{aligned}
   & \underset{B,U}{\mathop{\min }}\,{{\left\| B-UH \right\|}^{2}}  \\
   & s.t.~~B\in {{\left\{ -1,1 \right\}}^{c\times n}}.\\
\end{aligned}
\end{equation}
To take the advantage of label information to enhance the performance, we further introduce the linear regression with the label information and binary code in the objective function. The rational of the objective function is that good binary codes should be able to have minimum regression loss. Then we have:
\begin{equation}
\label{eq_18}
\begin{aligned}
  & \underset{W,B}{\mathop{\min }}\, {{\left\| Y-{{W}^{T}}B \right\|}^{2}}+\lambda {{\left\| W \right\|}^{2}} \\ 
 & \begin{matrix}
   s.t. & B\in {{\left\{ -1,1 \right\}}^{c\times n}},
\end{matrix} \\ 
\end{aligned}
\end{equation}
where $W$ is the classification projection matrix, $Y=\left\{{{y}_{1}},{{y}_{1}}\cdots {{y}_{n}} \right\}\in {{R}^{l\times n}}$ is labels matrix consisting of label vectors of pictures and $l$ is the length of label vector. 
By combining the equations Eqn.(\ref{eq_17}) and Eqn.(\ref{eq_18}), the final objective function of our BSDH is formulated as
\begin{equation}
\label{eq_19}
\begin{aligned}
  & \underset{B,W,U}{\mathop{\min }}\,{{\left\| Y-{{W}^{T}}B \right\|}^{2}}+\lambda {{\left\| W \right\|}^{2}}+\mu {{\left\| B-UH \right\|}^{2}} \\ 
 & s.t.~~B\in {{\left\{ -1,1 \right\}}^{c\times n}},   \\
\end{aligned}
\end{equation}
where $ H= [\operatorname{vec}( Q_{1}^{T}{{X}_{1}}{{Q}_{2}}), ...,\operatorname{vec}( Q_{1}^{T}{{X}_{n}}{{Q}_{2}})]$.

\section{Optimization Algorithm}
There are three variables to be updated in solving problem (\ref{eq_19}). The most challenging task comes from the binary constraint of $B$, which leads to an NP-hard problem. To this end, the existing methods directly drop the binary constraint or substitute it with continuous functions. However, both strategies may cause large quantity loss \cite{liu2017}, resulting in degraded performance. Based on recent research results \cite{shen_sdh,cui2018}, we propose a discrete optimization to address the resulting optimization problem (\ref{eq_19}) instead of using any relaxing scheme. We employ an alternating optimization algorithm to iteratively update variables, i.e. $B$, $W$ and $U$. That is, we optimize each variable when fixing others.

\textbf{Updating $W$ with others fixed:} Taking $W$ as the variable while the other variables being fixed, the sub-objective function to solve this problem is:
\begin{equation}
\label{eq_20}
\begin{aligned}
  & \underset{B,W,U}{\mathop{\min }}\,{{\left\| Y-{{W}^{T}}B \right\|}^{2}}+\lambda {{\left\| W \right\|}^{2}}\\ 
 \Leftrightarrow &\underset{B}{\mathop{\min}} \operatorname{tr}({{Y}^{T}}Y)-\operatorname{tr}({{B}^{T}}WY)-\operatorname{tr}({{Y}^{T}}{{W}^{T}}B)\\
 & ~~~~~+\operatorname{tr}({{B}^{T}}W{{W}^{T}}B)+\lambda \operatorname{tr}({{W}^{T}}W).
\end{aligned}
\end{equation}
The first term in the above equation is constant. Considering the function ${{N}}(W)$ which consists of the last four terms:
\begin{equation}
\label{eq_21}
\begin{aligned}
{N}(W) &=-\operatorname{tr}({{B}^{T}}WY)-\operatorname{tr}({{Y}^{T}}{{W}^{T}}B)\\
& +\operatorname{tr}({{B}^{T}}W{{W}^{T}}B)+\lambda \operatorname{tr}({{W}^{T}}W).
\end{aligned}
\end{equation}
Then we aim to 
\begin{equation}
\label{eq_22}
    \underset{W}{\mathop{\min}}\,{N}(W)=\underset{W}{\mathop{\min }}\, \begin{aligned}
  & \operatorname{tr}({{B}^{T}}WY)-\operatorname{tr}({{Y}^{T}}{{W}^{T}}B) \\ 
 & +\operatorname{tr}({{B}^{T}}W{{W}^{T}}B)+\lambda \operatorname{tr}({{W}^{T}}W). \\ 
\end{aligned}
\end{equation}
To solve problem (\ref{eq_22}), $W$ should satisfy the following condition:
\begin{equation}
\label{eq_23}
(B{{B}^{T}}+\lambda I)W=B{{Y}^{T}}.
\end{equation}
We can finally obtain $W$:
\begin{equation}
\label{eq_24}
W={{(B{{B}^{T}}+\lambda I)}^{-1}}B{{Y}^{T}}.
\end{equation}

\par \textbf{Updating $U$ with others fixed:}
The objective function with respect to $U$ is:
\begin{equation}
\label{eq_25}
\underset{U}{\mathop{\min }}\, {{\left\| B-UH \right\|}^{2}}.
\end{equation}         
Take the derivation with respect to $U$ and make it equal to zero, we can obtain the solution as follows:
\begin{equation}
\label{eq_26}
U=B{{H}^{T}}{{\left( H{{H}^{T}} \right)}^{-1}}.
\end{equation}

\par \textbf{Update $B$ with others fixed:} Updating $B$ is the most challenging part for the whole optimizing process because we need to handle the binary constraints, i.e., the sgn function in this part. The sub-objective function for $B$ is
\begin{equation}
\label{eq_27}
  \begin{aligned}
  & \underset{B}{\mathop{\min }}\, {{\left\| Y-{{W}^{T}}B \right\|}^{2}}+\upsilon {{\left\| B-UH \right\|}^{2}}\\ 
 & \begin{matrix}
   s.t. & B\in {{\left\{ -1,1 \right\}}^{c\times n}}.  \\
\end{matrix} \\ 
\end{aligned}
\end{equation}
We have the following optimization problem from Eqn.(\ref{eq_27}):
\begin{equation}\small
\label{eq_28}
\begin{aligned}
  &\underset{B}{\mathop{\min }}\, \operatorname{tr}({{(Y-{{W}^{T}}B)}^{T}}(Y-{{W}^{T}}B))\\ 
  &+\upsilon \operatorname{tr}({{(B-UH)}^{T}}(B-UH))\\ 
 & s.t.~B\in {{\left\{ -1,1 \right\}}^{c\times n}}, \\ 
\end{aligned}
\end{equation}
\begin{equation}
\label{eq_29}
\begin{aligned}
  \Leftrightarrow &\underset{B}{\mathop{\min }}\, \operatorname{tr}({{B}^{T}}W{{W}^{T}}B)-2\operatorname{tr}({{B}^{T}}WY)\\
  &-2\upsilon \operatorname{tr}({{H}^{T}}{{U}^{T}}B) \\ 
 & s.t.~B\in {{\left\{ -1,1 \right\}}^{c\times n}}. \\ 
\end{aligned}
\end{equation}
It is noted that tr$({{B}^{T}}B)=cn$ is constant and thus the formulation turned into:
\begin{equation}
\label{eq_30}
\begin{aligned}
  & \underset{B}{\mathop{\min }}\,{{\left\| {{W}^{T}}B \right\|}^{2}}-2\operatorname{tr}({{B}^{T}}WY)-2\upsilon \operatorname{tr}({{H}^{T}}{{U}^{T}}B)\\ 
 & s.t.~B\in {{\left\{ -1,1 \right\}}^{c\times n}}. \\ 
\end{aligned}
\end{equation}
To solve problem (\ref{eq_30}), we use discrete cyclic coordinate (DCC) descent method to learn $B$ row by row, i.e., to learn ${{b}_{i}}$ bit by bit. Let ${{v}^{T}}$ be ${{l}^{th}}$ row of $W$ , ${W}'$ be the rest matrix of $W$, and ${{z}^{T}}$ be ${{l}^{th}}$ row of $B$, ${B}'$ be the rest matrix of $B$. Now we consider the first term of (\ref{eq_30}):
\begin{equation}
\label{eq_31}
\begin{aligned}
  & {{\left\| {{W}^{T}}B \right\|}^{2}}={{\left\| {{\left( \begin{matrix}
   {{W}'}  \\
   {{v}^{T}}  \\
\end{matrix} \right)}^{T}}\left( \begin{matrix}
   {{B}'}  \\
   {{z}^{T}}  \\
\end{matrix} \right) \right\|}^{2}}={{\left\| {{{{W}'}}^{T}}{B}'+v{{z}^{T}} \right\|}^{2}} \\ 
 & ={{\left\| {{{{W}'}}^{T}}{B}' \right\|}^{2}}+2\operatorname{tr}\left( z{{v}^{T}}{{{{W}'}}^{T}}{B}' \right)+\operatorname{tr}\left( z{{v}^{T}}v{{z}^{T}} \right).  
\end{aligned}
\end{equation}
When ${W}'$, ${B}'$ are fixed and $z$ is variable, for Eqn.(\ref{eq_31}) we have:
\begin{equation}
\label{eq_32}
{{\left\| {{{{W}'}}^{T}}{B}' \right\|}^{2}}=const,
\end{equation}
\begin{equation}\small
\label{eq_33}
\begin{aligned}
\operatorname{tr}\left( z{{v}^{T}}v{{z}^{T}} \right)&=\operatorname{tr}\left( {{z}^{T}}z{{v}^{T}}v \right)
=\operatorname{tr}\left(n{{v}^{T}}v \right)=n\operatorname{tr}\left( {{v}^{T}}v \right)=const,
\end{aligned}
\end{equation}
where $const$ denotes constant. Thus, only second term in Eqn.(\ref{eq_31}) is variable, and Eqn.(\ref{eq_31}) turns into:
\begin{equation}
\label{eq_34}
{{\left\| {{W}^{T}}B \right\|}^{2}}=const+2\operatorname{tr}\left( z{{v}^{T}}{{{{W}'}}^{T}}{B}' \right).
\end{equation}
Now we consider the second and third term of Eqn.(\ref{eq_30}):
\begin{equation}
\label{eq_35}
\begin{aligned}
  & -2\operatorname{tr}\left( {{B}^{T}}WY \right)-2\upsilon \operatorname{tr}\left( {{H}^{T}}{{U}^{T}}B \right) \\ 
  = & -2\operatorname{tr}\left( \left( {{Y}^{T}}{{W}^{T}}+\upsilon {{H}^{T}}{{U}^{T}} \right)B \right).  
\end{aligned}
\end{equation}
Let $M={{Y}^{T}}{{W}^{T}}+\upsilon {{H}^{T}}{{U}^{T}}$, $m$ be the ${{l}^{th}}$ column of $M$ and ${M}'$ be the rest matrix of $M$, Eqn.(\ref{eq_35}) can be reformed to:
\begin{equation}
\label{eq_36}
\begin{aligned}
-2\operatorname{tr}\left( MB \right)&=-2\operatorname{tr}\left( {M}'{B}'+m{{z}^{T}} \right)\\
&=const-2\operatorname{tr}\left( m{{z}^{T}} \right).
\end{aligned}
\end{equation}
After above derivations, the objective function (\ref{eq_30}) turns into:
\begin{equation}
\label{eq_37}
\underset{z}{\mathop{\min }}\,\left( {{v}^{T}}{{{{W}'}}^{T}}{B}'-{{m}^{T}} \right)z.     
\end{equation}
The optimal solution of problem (\ref{eq_37}) is:
\begin{equation}
\label{eq_38}
z= \operatorname{sgn} \left( {{m}^{T}}-{{v}^{T}}{{{{W}'}}^{T}}{B}' \right).
\end{equation}

In summary, the optimal solution is obtained by alternatively updating variables, and the whole optimization problem is divided into few steps. In the step of updating $B$, each bit is calculated based on the rest bits iteratively. The convergence of the proposed algorithm is promising. Experimental results show that the objective function value can be convergent within ten iterations. The detailed algorithm steps are presented in algorithm 1.

\begin{table}[!t] 
 \begin{tabular}{m{8cm}}
  \toprule 
  Algorithm 1. Learning Algorithm for BSDH \\ 
  \midrule
	Input: training data $\left\{ {{X}_{1}},{{X}_{2}},\ldots {{X}_{n}} \right\}$ ; label matrix $Y$; binary code length $c$; Transition sizes ${{c}_{1}}$ and ${{c}_{2}}$. maximum iteration number ${{t}_{1}}$, ${{t}_{2}}$; hyper-parameters $\lambda $, $\mu $;\\
	
    Output: binary codes $B$; classification projection matrix $W$, bilinear projection matrices ${{Q}_{1}}$ and ${{Q}_{2}}$.\\
  
1.	Compute the mean of dataset ${{M}_{0}}$ and the mean of $i\text{th}$  class${{M}_{i}}$. Initialize bilinear projection matrix ${{Q}_{2}}$ as identity matrix.\\

2.	Iteratively update ${{Q}_{2}}$ and ${{Q}_{1}}$ iteratively by solving the eigen decomposition problem of ${{\left( S_{w}^{{{Q}_{2}}} \right)}^{-1}}S_{B}^{{{Q}_{2}}}$ and ${{\left( S_{w}^{{{Q}_{1}}} \right)}^{-1}}S_{B}^{{{Q}_{1}}}$, repeat for ${{t}_{1}}$ times.\\

3.	Calculate $H$ by equation (15) and initialize $B$ as ${{\left\{ -1,1 \right\}}^{c\times n}}$  randomly.\\

4.	Repeat the following steps until converge or reach maximum iterations ${{t}_{2}}$:
\begin{itemize}  
\item calculate $W$ by equation (24).
\item compute  $U$ by equation (26).
\item learn $B$ bit by bit using DCC descent method by equation (38).
\end{itemize} \\
  \bottomrule 
 \end{tabular} 
\end{table}

\section{Experiments}
To evaluate the performance of the proposed BSDH method for image retrieval task, extensive experiments were conducted on four image datasets using different input features. The experimental results of BSDH were further compared with the results of existing hashing baselines, including traditional hashing methods and deep hashing methods. 
\subsection{\textbf{Datasets and Baselines}}
\subsubsection{\textbf{Datasets}}
We evaluate the performance of the proposed method on four benchmark datasets, i.e.  MNIST \cite{lecun1998}, CIFAR-10 \cite{krizhevsky2009}, PASCAL\_VOC and MirFlickr \cite{huiskes}. Specifically, images in MNIST and CIFAR-10 only have single label while images in the other two datasets have multiple labels. The ground-truth is defined based on labels, that  is, images sharing at least one label are considered as similar. More detailed introduction of these datasets are as follows.
\par \textbf{MNIST} consists of 60,000 $28\times 28$ small grey-scale images of handwritten digits from ‘0’ to ‘9’ in the training set. We randomly chose 54,000 images as the training set and used the rest 6000 images for testing.  For our method and other bilinear projection methods, the raw pixel matrix was used as input data. For other methods, the pixel matrix was vectorized to a 784-dimensional vector as the input.
\par \textbf{PASCAL\_VOC} is offered for annual PASCAL\_VOC challenge of image classification, detection and segmentation containing similar size images. We used VOC2012 in this paper and randomly chose 15,000 images of 20 classes in our experiment. All images were divided into two parts, i.e. 10k for training and 1500 for testing.  
\par \textbf{MirFlickr} is a micro version of dataset MirFlickr-1M. This dataset consists of 25,000 images with 24 classes of objects. In our experiment, we labeled this dataset based on its original annotations. Specifically, each image in the dataset was labeled with a 24-dimensional vector corresponding to 24 classes of objects. For image allocation, we used 10000 images to train our hashing coding method and another 5000 images to test the hashing model. 
\par \textbf{CIFAR-10} consists of 60,000 $32\times 32$ tiny color images from 10 categories, each of which has 6,000 samples. We randomly selected 1,000 images from each category and constructed a test set with 10,000 images. The remaining 50,000 images were used as training data. The CIFAR-10 dataset was used for conducting comprehensive comparative experiments to evaluate the effectiveness of the proposed BSDH method, especially compared to some deep baselines and non-deep methods using deep features. 
\subsubsection{\textbf{Baselines}}
To demonstrate the effectiveness of the proposed method, our BSDH is compared against classic or state-of-the-art competing hashing schemes, including $11$ non-deep hashing methods and $11$ deep hashing methods. Specifically, non-deep hashing methods include ITQ \cite{gong_itq}, SKLSH \cite{raginsky_locality}, SH \cite{weiss_sh}, DSH \cite{jin2014}, PCAH \cite{wang2012}, LSH \cite{dataracm}, BDAH \cite{liu2016}, KSH \cite{liu_ksh}, LFH \cite{zhang_supervised}, BRE[8] and MLH \cite{norouzi_minimal}, and deep hashing methods are CNNH \cite{xia_supervised}, CNNH+ \cite{xia_supervised}, DNNH \cite{lai2015}, DHN \cite{zhu_deephashing}, DSH(2)\cite{liu_fastdeephashing}, DQN \cite{cao_dqn}, VDSH \cite{zhang_efficient}, DPSH \cite{li2015}, DSRH \cite{zhao_deepsemantic}, DSCH \cite{zhang2015}, DRSCH \cite{zhang2015}. 

There are two hyper-parameters in the proposed algorithm, i.e., $\lambda $ and $\mu $. We employ the grid-search strategy to find the best parameters from the set $\{10^{-5}, 10^{-4}, 10^{-3}, 10^{-2}, 10^{-1}, 1, 10\}$. In the experiments, we set $\lambda  = {{10}^{-5}}$ and $\mu = {{10}^{-1}}$ for our method on all the datasets. For most baseline methods, we re-implemented the source codes provided by authors with default parameters. For those methods who unreleased codes such as BDAH, we wrote the codes by ourselves according to the algorithm stated in original papers. To make fair comparison, all experiments were conducted five times for all methods, and the averaged performance are reported.

\subsection{\textbf{Experimental Setting}}	
\subsubsection{\textbf{Image feature extraction}}
An effective image feature extraction algorithm is foundation of many image analysis tasks. To incorporate proper image features into the proposed BSDH method and further investigate the influence of input features, we design two 2D feature extraction methods: a traditional one based on SURF [28] and a deep feature based on CNN \cite{krizhevsky} , which are introduced below in detail . 
\par \textbf{Traditional 2D feature - FVLAD:} As we know, VLAD [13] feature is an effective 2D feature which is detectable even under changes in image scale, noise and illumination. It is constructed based on the SIFT (scale invariant feature transform) \cite{lowe2004}. However, SIFT has the obvious drawback for its application, i.e., computational complexity. To avoid heavy computation, we design a descriptor based on SURF (speed up robust feature), which is much faster to calculate, almost three times faster than SIFT \cite{bauer2007}. The new descriptor is similarly constructed as VLAD and named as FVLAD (Fast VLAD). \par \textbf{Deep 2D feature - AlexConv5:} There is no doubt that convolutional neural network (CNN) has been the most powerful tool in computer vision. To further testify the effectiveness of our method, we designed a new 2D convolutional feature and evaluated the performance of the proposed method with this feature. Notably, the most widely used CNN features are extracted from the fully-connected (FC) layer \cite{donahueicml}. However, such features cannot capture the spatial information, leading to inferior performance. Therefore, we chose to directly take the output of the convolutional layers as the input feature for the proposed BSDH method. Specifically, we adopted the AlexNet \cite{krizhevsky} pre-trained on ImageNet dataset as the CNN feature extractor and extracted the output of the last convolutional layer (5th conv layer) as the image feature, which is with the size of $256\times 36$ and named as AlexConv5. 

\subsubsection{\textbf{Evaluation protocols}} When binary codes learning was finished, hamming distance was calculated to measure the difference between the query images and training images, and images with smaller difference to queries will be retrieved as target results. To evaluate the image retrieval performance, the commonly used mean average precision (MAP) \cite{mcsherry_computing}, recall rate with the number of retrieved samples, precision rate with the number of retrieved sample and precision-recall curves are used as the criteria for performance evaluation. Considering the result vector 
\begin{equation}
V=\left\{ {{v}_{1}},{{v}_{2}},\ldots ,{{v}_{n}} \right\},
\end{equation}
for a query $q$ after hash ranking, the recall rate, precision rate and MAP are defined as follows \cite{mcsherry_computing}: 
\begin{equation}
P_{k}\left( V \right)=\frac{1}{k}\sum\limits_{i=1}^{k}{rel\left( {{v}_{i}} \right)},
\end{equation}
\begin{equation}
R_{k}\left( V \right)=\frac{\sum\nolimits_{i=1}^{k}{rel\left( {{v}_{i}} \right)}}{\sum\nolimits_{i=1}^{n}{rel\left( {{v}_{i}} \right)}},
\end{equation}
\begin{equation}
MAP=mean\left( \frac{\sum\nolimits_{i=1}^{k}{P_{i}\left( V \right)rel\left( {{v}_{i}} \right)}}{\sum\nolimits_{i=1}^{n}{rel\left( {{v}_{i}} \right)}} \right),
\end{equation}
where $rel\left( {{v}_{i}} \right)=\left\{ \begin{matrix}
   1 & relevant  \\
   0 & irrelevant  \\
\end{matrix} \right.$ denotes the relevance of the ${{v}_{i}}$. 

\subsection{\textbf{Experimental Results and Analysis}}
To comprehensively evaluate the performance of the proposed method, we conducted extensive experiments with different input features on different datasets. The detailed experimental layout is illustrated in Table  \uppercase\expandafter{\romannumeral2}.

\begin{table}[!t]
\caption{Input feature setting for different datasets}
\begin{center}
\begin{tabular}{m{1.8cm}<{\centering} m{1.8cm}<{\centering} m{2cm}<{\centering}}
\hline
{Datasets}&{Feature type}&{Feature size} \\
\cline{1-3} 
MNIST  &Raw pixel &28 $\times$ 28\\
PASCAL\_VOC  &FVLAD &100 $\times$ 64\\
MirFlickr &FVLAD &100 $\times$ 64 \\
CIFAR-10 &AlexConv5 &256 $\times$ 36 \\
\hline
\end{tabular}
\label{tab2}
\end{center}
\end{table}

\subsubsection{\textbf{Experimental results on MNIST (raw pixel features)}}
The MAP result, recall rate curve with retrieved number, precision rate curve with retrieved number and precision-recall curve of the proposed method and baselines on MNIST are reported in this section. MAP results are shown in Table  \uppercase\expandafter{\romannumeral3}. 

\begin{table}[!t]
\caption{ MAP result on MNIST}
\begin{center}
\begin{tabular}{m{1.5cm}<{\centering} m{1.0cm}<{\centering} m{1.0cm}<{\centering} m{1.0cm}<{\centering} m{1.0cm}<{\centering}}
\hline
{Code Length}&{16}&{32}&{64}&{128} \\
\cline{1-5} 
{LSH}  &0.182 &0.263 &0.295 &0.326   \\
{PCAH}  &0.255 &0.224 &0.198  &0.174\\
{SH} &0.277 &0.264 &0.248  &0.260 \\
{DSH} &0.302 &0.314 &0.371  &0.390 \\
{ITQ}  &0.387 &0.427 &0.451  & 0.463 \\
{BDAH}  &0.483 &0.541 &0.568  &0.581   \\
{LFH}  &0.542 &0.541 &0.568  &0.581  \\
{KSH} &0.806 &0.795 &0.838  &0.834  \\
{BSDH} & \textbf{0.844}& \textbf{0.878}& \textbf{0.888} & \textbf{0.892}\\
\hline
\end{tabular}
\label{tab3}
\end{center}
\end{table}

\par It is clear in Table  \uppercase\expandafter{\romannumeral3} that our method outperforms all other compared method in MAP results. It is worth noting that among all comparison methods, BDAH also adopts the strategy of bilinear projection, but it only achieves the MAP around 0.5. With the similar idea of bilinear projection, the proposed BSDH method achieves the MAP over 0.8. The reason may be that our method integrates linear regression in the code learning and fully takes advantage of the label information. 
Further, compared with supervised hashing methods, namely BDAH, KSH, LFH, our method shows best performance in all configurations. As all compared methods learn hash codes under the supervision of label information, the main reason for BSDH obtaining better performance is probably because it directly conducts bilinear projection on pixel matrix and does not change it to vector manually. According to previous research, the matrix input contains abundant structural information about spatial relationship between pixels, which will be destroyed if treated as vectors [16]. 
\begin{figure*}[t]
\begin{center}
    \includegraphics[width=7in]{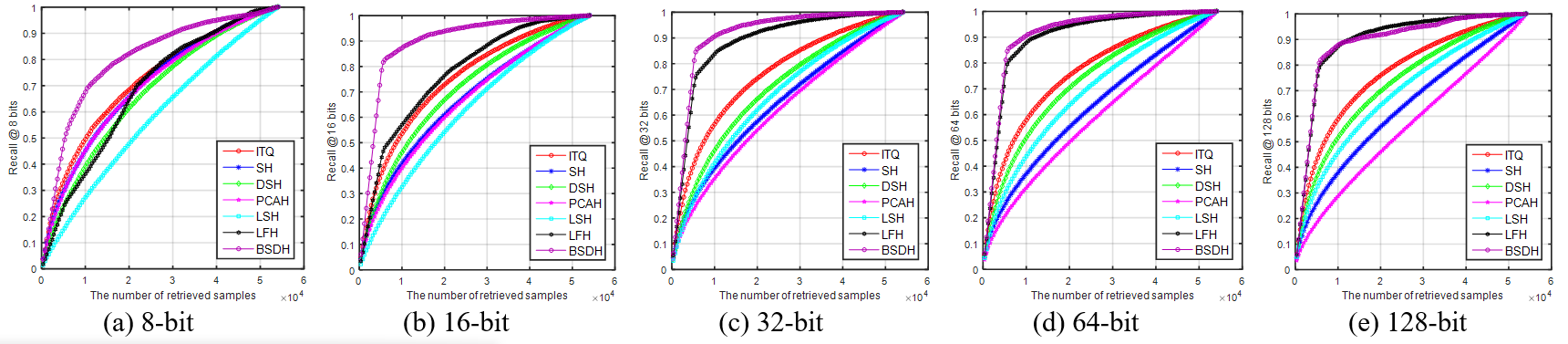}
    \caption{Recall curve on MNIST with code length from 8-bit to 128-bit.}
\end{center}
\label{fig2}
\end{figure*}
\begin{figure*}[t]
\begin{center}
    \includegraphics[width=7in]{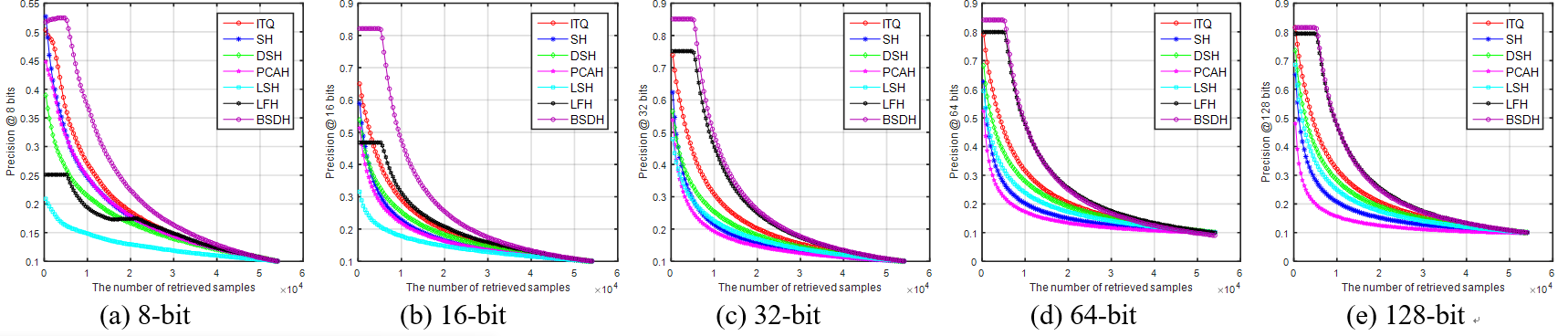}
    \caption{Precision curve on MNIST with code length from 8-bit to 128-bit.}
\end{center}
\label{fig3}
\end{figure*}
\begin{figure*}[t]
\begin{center}
    \includegraphics[width=7in]{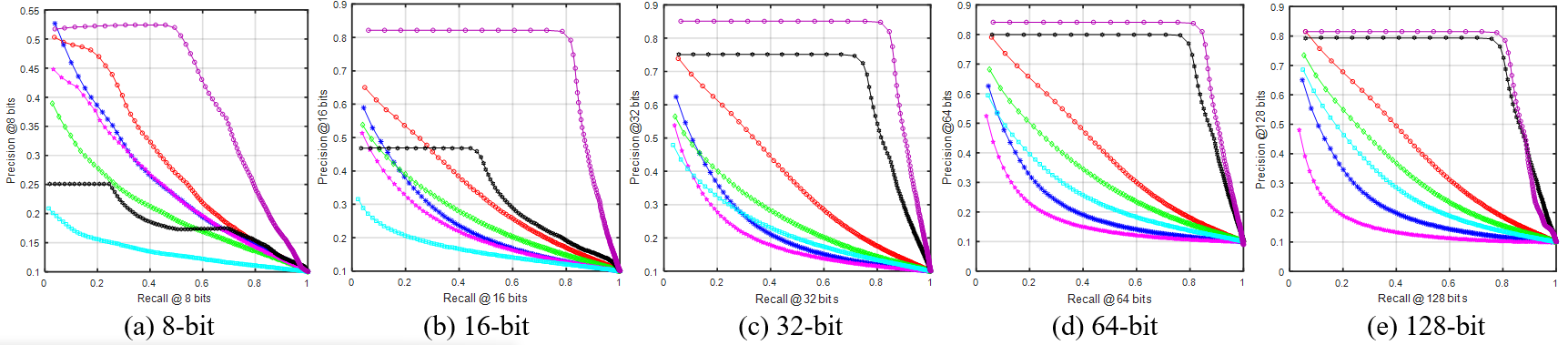}
    \caption{Recall-Precision curve on MNIST with code length from 8-bit to 128-bit.}
\end{center}
\label{fig4}
\end{figure*}
\par Figure 2-4 illustrate the recall rate, precision rate and recall-precision curve with respect to different code length from 8 to 128. It is clear that the BSDH method almost outperforms all compared methods on all three criterions, especially for recall-precision curves where it shows superior performance. It can also be seen that our method shows consistently good performance on the cases of different code lengths, except for very short code of 8-bits. Comparatively, one baseline method LFH can also achieve good results when the code length is long, while for short code cases, its performance is apparently degraded. 

\subsubsection{\textbf{Experimental results on  PASCAL and MirFlickr (FVLAD features)}}
\par Table  \uppercase\expandafter{\romannumeral4} and Table  \uppercase\expandafter{\romannumeral5} report the MAP results of comparison experiments on PASCAL and MirFlickr respectively. In this section, we compared our method with other seven methods, including five unsupervised methods and two supervised methods. From Table  \uppercase\expandafter{\romannumeral4}, we can see that BSDH still achieves good results on PASCAL. Specifically, it outperformed all other compared baselines in different code length and it is the only method achieved results over 0.3 on this dataset. It is worth noting that the overall performance on this dataset is not good. The possible reason may be that images in that dataset are not mostly correlative, which means that the similar pairs are fewer and the useful information for learning is limited. We also show the recall-precision results on PASCAL in Fig. 5. From the three figures, we can see that although BSDH does not achieve good performance when retrieving very few samples, it keeps the precision with the increasement of recall rate. This also explains why it can obtain better MAP results.
\begin{table}[htbp]
\caption{ MAP results on PASCAL}
\begin{center}
\begin{tabular}{m{1.7cm}<{\centering} m{1.2cm}<{\centering} m{1.0cm}<{\centering} m{1.2cm}<{\centering} m{1.2cm}<{\centering}}
\hline
{Code Length}&{16}&{32}&{64}&{128} \\
\cline{1-5} 
LSH &0.180  &0.180 &0.181 &0.181\\
SH	&0.188	&0.185	&0.183	&0.186\\
PCAH &0.184	&0.182	&0.181	&0.179\\
DSH	&0.183	&0.188	&0.191	&0.189\\
ITQ	&0.189	&0.187	&0.184	&0.181\\
BDAH &0.193	&0.194	&0.194	&0.199\\
KSH	&0.248	&0.194	&0.184	&0.184\\
LFH	&0.273	&0.271	&0.273	&0.262\\
{BSDH} & \textbf{0.363}& \textbf{0.360}& \textbf{0.363} & \textbf{0.384}\\
\hline
\end{tabular}
\label{tab4}
\end{center}
\end{table}

\begin{figure*}[!t]
  \centering
  \subfigure[Recall-Precision curve of 32-bit codes]{
    \label{fig:subfig:a} 
    \includegraphics[width = 2.0in]{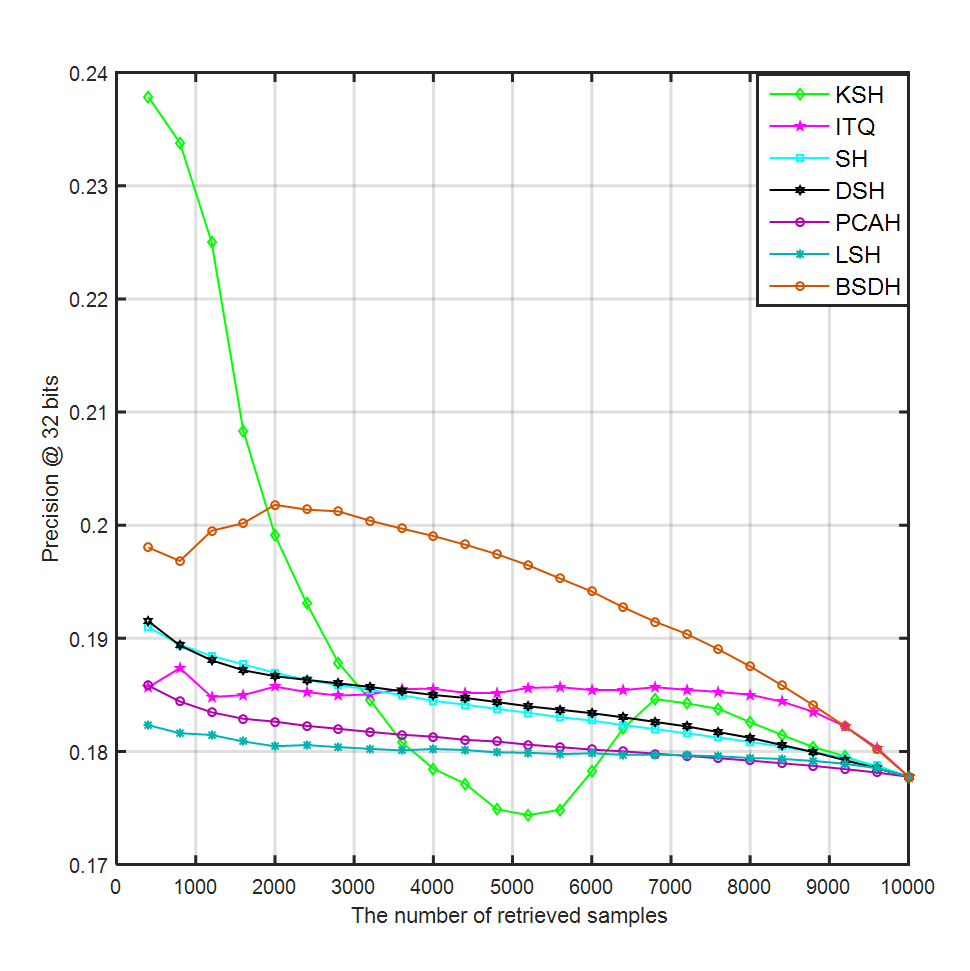}}
  \subfigure[Recall-Precision curve of 64-bit codes]{
    \label{fig:subfig:b}
    \includegraphics[width = 2.0in]{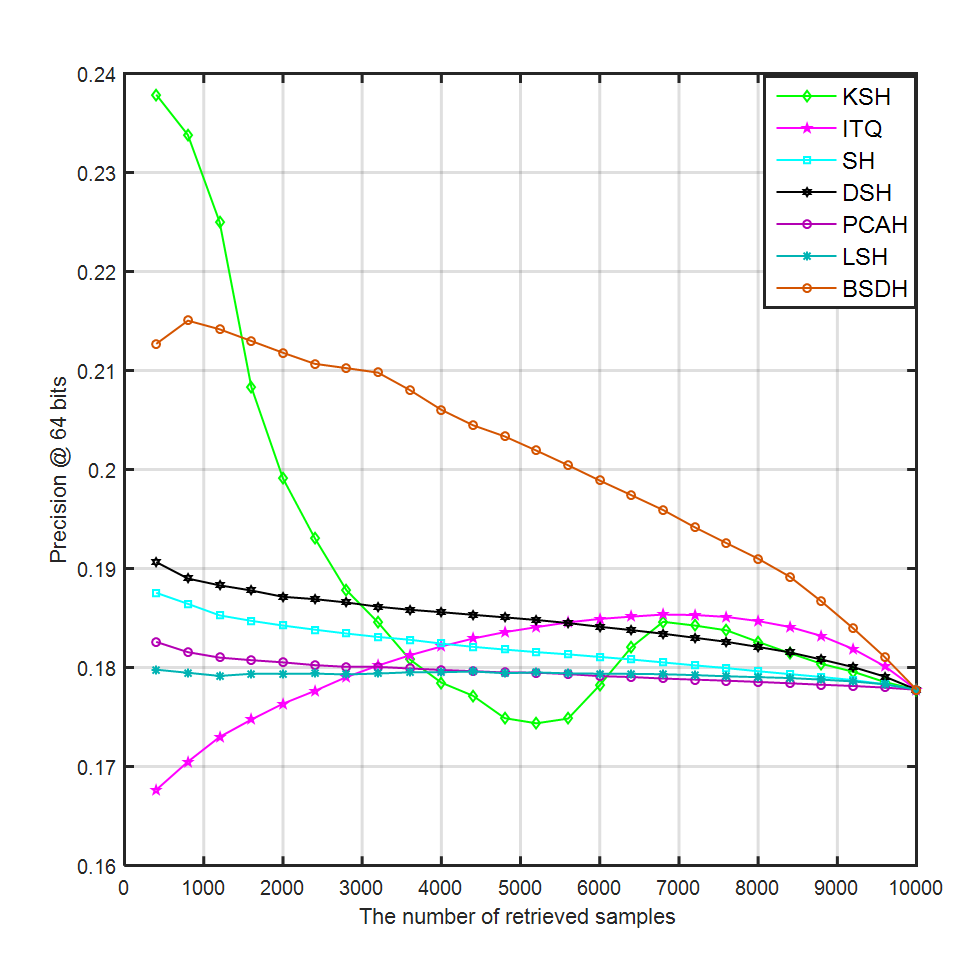}}
  \subfigure[Recall-Precision curve of 128-bit codes]{
    \label{fig:subfig:b}
    \includegraphics[width = 2.0in]{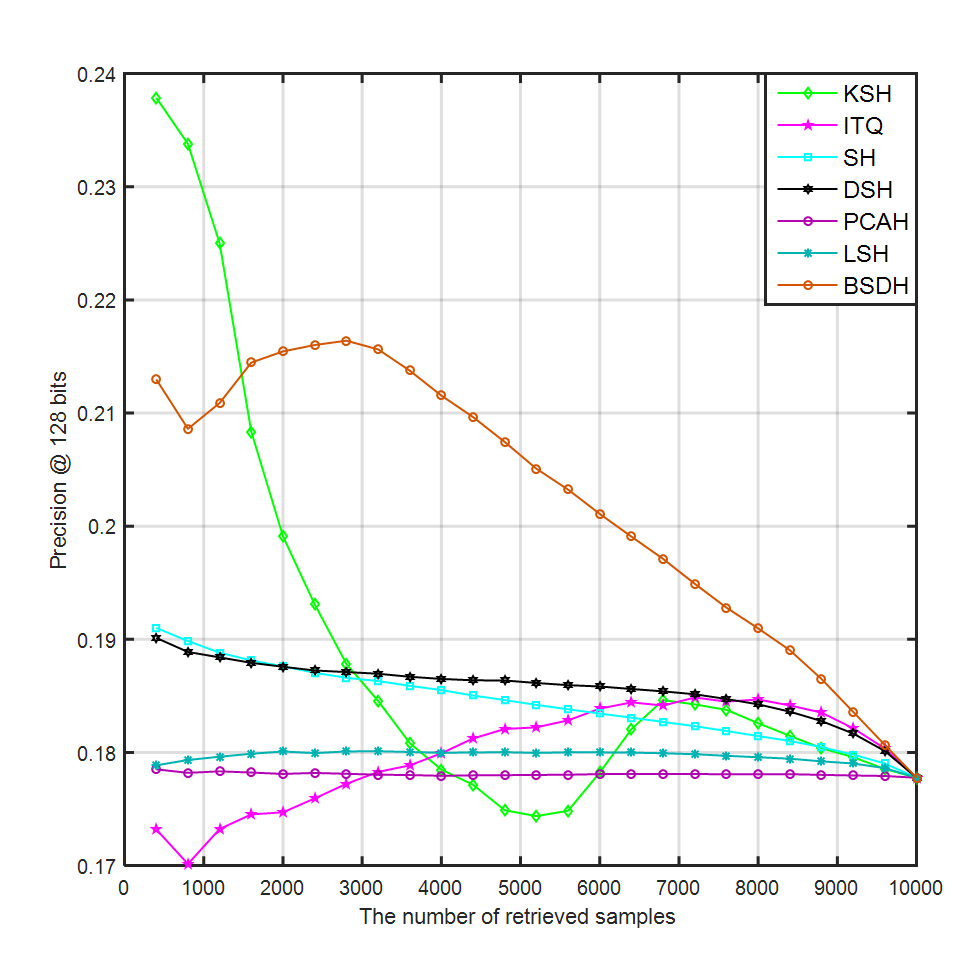}}
  \caption{Recall-Precision curve on PASCAL with respect to various code length}
  \label{fig6} 
\end{figure*}

\par From Table  \uppercase\expandafter{\romannumeral5}, it is clear that our BSDH outperforms all the compared methods in all cases, which further demonstrates the superiority of the proposed bilinear learning framework. Another observation from Table  \uppercase\expandafter{\romannumeral5} is that the overall results are much better than that of PASCAL. This is because the training images we used are labeled by rich information which makes the relative targets easier to recall.
\begin{table}[htbp]
\caption{ MAP results on MirFlickr}
\begin{center}
\begin{tabular}{m{1.7cm}<{\centering} m{1.2cm}<{\centering} m{1.2cm}<{\centering} m{1.2cm}<{\centering} m{1.2cm}<{\centering}}
\hline
{Code Length}&{16}&{32}&{64}&{128} \\
\cline{1-5} 
LSH &0.543	&0.545	&0.546	&0.547\\
SH	&0.552	&0.551	&0.548	&0.548\\
PCAH	&0.546	&0.546	&0.545	&0.545\\
DSH	&0.551	&0.552	&0.555	&0.556\\
ITQ	&0.539	&0.541	&0.544	&0.549\\
BDAH	&0.551	&0.553	&0.553	&0.552\\
KSH	&0.544	&0.545	&0.546	&0.548\\
LFH	&0.578	&0.579	&0.582	&0.585\\
{BSDH} & \textbf{0.635}& \textbf{0.656}& \textbf{0.672} & \textbf{0.689}\\
\hline
\end{tabular}
\label{tab5}
\end{center}
\end{table}
\par We used the FVLAD feature as the inputs on both PASCAL and MirFlickr for the proposed BSDH and other baselines. The results of these datasets show that this feature can help achieve good retrieval performance, especially when integrating with our bilinear projection strategy. To further support this point, we conducted extra experiments on PASCAL to evaluate the effectiveness of the used FVLAD feature. Since FVLAD can be seen as a fast version of VLAD, we attracted VLAD feature on the database and conducted image retrieval experiments using three hashing methods (including the proposed BSDH method). The experimental results are shown in Table \uppercase\expandafter{\romannumeral6}. Results in Table  \uppercase\expandafter{\romannumeral6} clearly show that the proposed FVLAD feature is very solid to be used as low-level image feature for hashing algorithms. 
\begin{table}[hbp]
\caption{MAP results for image retrieval using FVLAD and VLAD features on PASCAL by different methods}
\begin{center}
\begin{tabular}{m{1.7cm}<{\centering} m{1.2cm}<{\centering} m{1.2cm}<{\centering} m{1.2cm}<{\centering} m{1.2cm}<{\centering}}
\hline
\multirow{2}*{Methods} &\multicolumn{2}{c}{16}&\multicolumn{2}{c}{32}\\
\cline{2-5} 
~ &VLAD	&\textbf{FVLAD}	&VLAD	&\textbf{FVLAD}\\
\cline{1-5} 
SH	&0.180	&\textbf{0.185}	&0.181	&\textbf{0.183}\\
DSH	&0.179	&\textbf{0.188}	&0.181	&\textbf{0.191}\\
BSDH	&0.311	&\textbf{0.360}	&0.340	&\textbf{0.363}\\

\hline
\end{tabular}
\label{tab6}
\end{center}
\end{table}

\subsubsection{\textbf{Experimental results on CIFAR-10 (CNN-based features)}}
In this part, we discuss the performance of the proposed BSDH method with deep features by comparing it with deep hashing methods and traditional hashing methods using deep features. Most compared data in this part are from published papers and will be noted clearly in proper place. As mentioned, the CNN-based image feature, AlexConv5, is firstly extracted and then fed into BSDH algorithm for conducting standard image retrieval experiments. The MAP is also adopted in this part as evaluation criteria. 
\begin{table}[htbp]
\caption{MAP results of BSDH with AlexConv5 features and traditional hashing baselines with CNN features on CIFAR-10}
\begin{center}
\begin{tabular}{m{1.7cm}<{\centering} m{1.2cm}<{\centering} m{1.2cm}<{\centering} m{1.2cm}<{\centering} }
\hline
{Code Length}&{12}&{24}&{48} \\
\cline{1-4} 
LSH	&0.131	&0.162	&0.203\\
ITQ	&0.271	&0.283	&0.291\\
PCAH	&0.206	&0.187	&0.170\\
SH	&0.194	&0.192	&0.175\\
LDAH	&0.219	&0.179	&0.159\\
BRE	&0.255	&0.267	&0.286\\
MLH	&0.174	&0.168	&0.174\\
KSH	&0.344	&0.462	&0.548\\
BSDH	&\textbf{0.474}	&\textbf{0.756}	&\textbf{0.773}\\
\hline
\end{tabular}
\label{tab7}
\end{center}
\end{table}
\par Firstly, we compared our BSDH method with traditional hashing methods in terms of the image retrieval performance with three lengths of hash code. All compared hashing methods adopted the 1024-d deep feature \cite{mu_deephashing}. Similar to our AlexConv5 feature, this deep feature was also extracted from the last convolutional layer of CNN, but the network structures were slightly different in terms of the kernel size and other details. This feature was in vector-form and related detailed information can be found in Mu's paper \cite{mu_deephashing}. MAP results of BSDH and baselines are listed in Table  \uppercase\expandafter{\romannumeral7}, and most results of baselines are cited from the paper \cite{mu_deephashing}. For our BSDH method, We conducted experiments with the same setting as Mu's paper \cite{mu_deephashing} except using different  deep features. 
\par From Table  \uppercase\expandafter{\romannumeral7}, we can see that the performances of most baselines do not improve much even when adopting deep image features. The MAPs for most methods are below 0.3 except for KSH, whose performance is better than others. However, the MAP results of KSH are still lower than 0.6. Compared with these baselines, the proposed BSDH method achieves markedly better MAP results. The MAPs of both 24-bit and 48-bit cases are more than 0.75, which are quite satisfactory. For 12-bit case, the result is not as good as the other two cases, but it is still the best compared with the other methods.
\begin{table}[htbp]
\caption{MAP results of BSDH and traditional hashing baselines both with AlexConv5 features on CIFAR-10}
\begin{center}
\begin{tabular}{m{1.7cm}<{\centering} m{1.2cm}<{\centering} m{1.2cm}<{\centering} m{1.2cm}<{\centering}}
\hline
{Code Length}&{12}&{24}&{48} \\
\cline{1-4} 
LSH	&0.114	&0.115	&0.119\\
ITQ &0.160	&0.174	&0.178\\
PCAH	&0.147	&0.140	&0.132\\
SH	&0.150	&0.142	&0.138\\
KSH	&0.366	&0.413	&0.436\\
BSDH	&0.474	&0.756	&0.773\\
\hline
\end{tabular}
\label{tab8}
\end{center}
\end{table}
\par In order to eliminate the influence of different image features on comparison of our method and baselines, we applied the vectorized AlexConv5 feature as the input of five typical baselines and conducted additional experiments. Under this circumstance, the comparison between baselines and our method is completely fair. The experimental results are listed in Table  \uppercase\expandafter{\romannumeral8}. It is clear that our BSDH method is still far better than others under this experimental condition, which firmly validates its’ effectiveness and shows the advancement of the BSDH method. In addition, comparing results from Table  \uppercase\expandafter{\romannumeral8} and Table  \uppercase\expandafter{\romannumeral7} also show the influence of difference features on the performance of hashing methods. Generally, the change of input features does not affect much on compared baselines’ performance with respect to the MAP results.

\begin{table}[htbp]
\caption{MAP results of BSDH and traditional hashing baselines with handcraft features and CNN features on CIFAR-10}
\begin{center}
\begin{tabular}{m{1.7cm}<{\centering} m{1.2cm}<{\centering} m{1.2cm}<{\centering} m{1.2cm}<{\centering} m{1.2cm}<{\centering}}
\hline
\multirow{2}*{Methods} &\multicolumn{2}{c}{Handcraft feature}&\multicolumn{2}{c}{CNN feature}\\
\cline{2-5} 
 ~  &16 &32	&16	&32\\
\cline{1-5} 
LSH	&0.121	&0.138	&0.135	&0.175\\
ITQ	&0.152	&0.160	&0.275	&0.286\\
BRE	&0.130	&0.136	&0.263	&0.280\\
MLH	&0.137	&0.133	&0.181	&0.180\\
KSH	&0.219	&0.208	&0.395	&0.503\\
DeepHash	&0.216	&0.230	&0.547	&0.567\\
BSDH	&\textbf{0.278}	&\textbf{0.299}	&\textbf{0.689}	&\textbf{0.754}\\
\hline
\end{tabular}
\label{tab9}
\end{center}
\end{table}

\par In Table  \uppercase\expandafter{\romannumeral9}, more comprehensive comparisons are made between the proposed BSDH method and traditional baselines with both handcraft features and CNN-based features. It is noted that the handcraft feature used in our method is the gray pixel matrix and CNN feature is the proposed AlexConv5. For other compared methods, 800-d GIST feature and 1024-d CNN feature were adopted according to \cite{mu_deephashing}. All comparable data are directly cited from Mu's work \cite{mu_deephashing}. 
\par From Table  \uppercase\expandafter{\romannumeral9}, we have following observations. Firstly, our BSDH method outperforms all baselines with both non-deep and deep features, and with deep feature, the advantage of the proposed BSDH method is more remarkable. Secondly, the improvements on performance resulting from deep features varies with different methods. For example, LSH and MLH are not benefited much from deep feature while ITQ, BRE and KSH show clearly better performance by using deep features. Among all baselines, the biggest improvement from altering input features is shown in DeepHash. Comparatively, by using CNN feature, the proposed BSDH method achieves the most significant improvement on its retrieval performance. Specifically, for the 16-bit case, the MAP increases from 0.278 to 0.689 (by 0.411). For the 32-bit case, it dramatically increases by around 0.45. The results show that our BSDH method may be suitable to handle deep features and is promising to achieve desired performance with deep image features. This is a great advantage of the BSDH method as nowadays there already exist many well-trained deep network models, deep features of images are not difficult to obtain, which makes the proposed method more suitable in real-world application.
\par Related research has shown that deep hashing methods achieved notable advancement. However recently it can be found in research papers that traditional non-deep hashing methods could have even better performance with the help of deep learning-based image features compared with integrated end-to-end deep hashing methods. As the experimental results illustrated in this paper above have proved that CNN feature can help improve performance of traditional hashing, in the following, we investigate if the proposed BSDH method can outperform any deep hashing methods using deep features. In Table  \uppercase\expandafter{\romannumeral10}, we compare the performance of our method using AlexConv5 feature with 11 state-of-the-art end-to-end deep hashing methods. Considering the dramatic training time and computing resource cost of deep hashing methods, we did not reimplement those methods. The compared data in Table  \uppercase\expandafter{\romannumeral10} are partly from original papers of those methods and partly from reference papers [48, 51, 53]. It is noticed that for the first eight deep hashing methods, the image number of the training set is 5000 as usual set in deep learning methods with the consideration of the computing cost. The rest three have same quantity of training data, database data and test data as our BSDH method. 
\begin{table}[htbp]
\caption{MAP results of BSDH with AlexCon5 features and deep hashing baselines on CIFAR-10}
\begin{center}
\begin{tabular}{m{1.7cm}<{\centering} m{1.2cm}<{\centering} m{1.2cm}<{\centering} m{1.2cm}<{\centering} m{1.2cm}<{\centering}}
\hline
{Code Length}&{24}&{32}&{48}&{64} \\
\cline{1-5} 
CNNH	&0.511	&0.542	&0.522	&-\\
CNNH+	&0.521	&0.521	&0.532	&-\\
DNNH	&0.566	&0.558	&0.581	&-\\
DHN	&0.594	&0.603	&0.621	&-\\
DSH(2)	&0.651	&0.659	&0.662	&0.671\\
DQN	&0.558	&0.564	&0.580	&-\\
VDSH &0.541	&0.545	&0.548	&-\\
DPSH	&0.727	&0.732	&0.778	&0.776\\
DeepHash*	&0.729	&0.735	&0.741	&0.748\\
DSCH* &0.613	&0.617	&0.620	&0.624\\
DRSCH* &0.622	&0.629	&0.631	&0.633\\
BSDH*	&\textbf{0.756}	&\textbf{0.754}	&\textbf{0.773}	&\textbf{0.777}\\
\hline
\end{tabular}
\end{center}
\footnotesize
* denotes the experimental settings of the corresponding methods are exactly same with our method, including the quantities of training data, database data and test data.
\label{tab10}
\end{table}
\par From Table  \uppercase\expandafter{\romannumeral10}, it is clear that the proposed BSDH method outperforms all compared deep hashing baselines with different hash code lengths. It is generally believed that end-to-end deep hashing methods usually achieve better performance because of good nonlinear fitting capability of deep neural networks \cite{lin2015}. However, to obtain desired performance, hours or even days are required to train the deep hashing model. In contrast, the proposed BSDH just takes a few minutes to learn hash codes, whose efficiency is much better. Importantly, with such small computational cost, the learned hash codes by BSDH are with desired representative ability considering the retrieval performance shown in Table  \uppercase\expandafter{\romannumeral10}. In summary, compared with end-to-end deep hashing methods, the proposed BSDH method has the following advantages: 1) the training time is much less; 2) the input is flexible and desired performance can be achieved by using CNN features; 3) parameters in the model are much less such that time spent on parameter setting can be reduced. Comprehensively speaking, the proposed BSDH method shows advantages compared to existing deep hashing methods.  

\subsection{\textbf{Parameter Sensitivity Analysis}}
\subsubsection{\textbf{Discussion on hyper-parameters}}
There are two hyper-parameters in the proposed algorithm, i.e., $\lambda $ and $\mu $. To analyze how the two parameters will affect the performance of the algorithm, we conducted experiments with different sets of values and made comparison on their performance on MNIST dataset.  We tried 81 combinations and the results are shown in Fig. 6. As we can see, apparently the value of $\lambda $ has more influence on the results. When $\lambda $  is taken in the range of less than ${{10}^{-5}}$, the algorithm shows quite satisfactory performance. However, if $\lambda $ is relatively big, for instance, one or ten, the algorithm almost failed. Observing the objective function, we can see that $\lambda $ is the weight of the regularization part. When the weight is too big, the algorithm will be regularized so hard that its fitting ability is weaken. Therefore, $\lambda $ should be set comparatively small so that it can help prevent overfitting and at mean time not weaken overall performance of the algorithm. On the other hand, the proposed method is robust to the values of $\mu $, except when $\lambda $ is in the range of  $[{{10}^{-4}},{{10}^{-2}}]$. This means that the quantization loss part in our objective function have more impact on the results when  $\lambda \in [{{10}^{-4}},{{10}^{-2}}]$. Meanwhile, when $\lambda $  is too small or too large, the quantization loss part has less effect.

\begin{figure}[!t]
\centering
\includegraphics[scale=0.4]{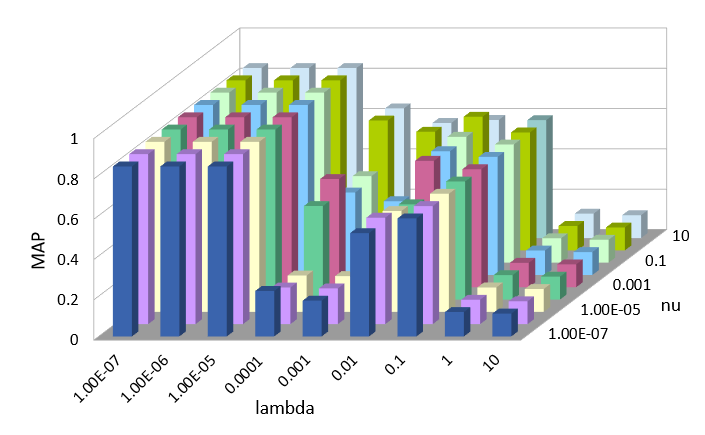}
\caption{Retrieval MAP on MNIST under different hyper-parameters. With $\lambda $ in the range of $\left[ {{10}^{-7}},10 \right]$, $\mu $  in the range of $\left[ {{10}^{-7}},10 \right]$.}
\label{fig6}
\end{figure}

\subsubsection{\textbf{Discussion on binary code length and transition size}}
In this part, we make discussion on the influence of the transition sizes (${{c}_{1}}$ and ${{c}_{2}}$) and the hash code lengths ($c$) on the performance of the proposed algorithm. We chose 12 groups of ${{c}_{1}}$ and ${{c}_{2}}$, and five code lengths to conduct image retrieval experiments on CIFAR-10, with the input feature as the proposed $256\times 36$ AlexConv5. In total, 60 sets of MAP results are illustrated in Fig. 7. From Fig 7, we can see that the performance is generally growing with the increasement of code length. The improvement is remarkable from 12-bit to 16-bit and becomes small for hash code more than 16-bit.
\begin{figure}[!t]
\centering
\includegraphics[scale=0.4]{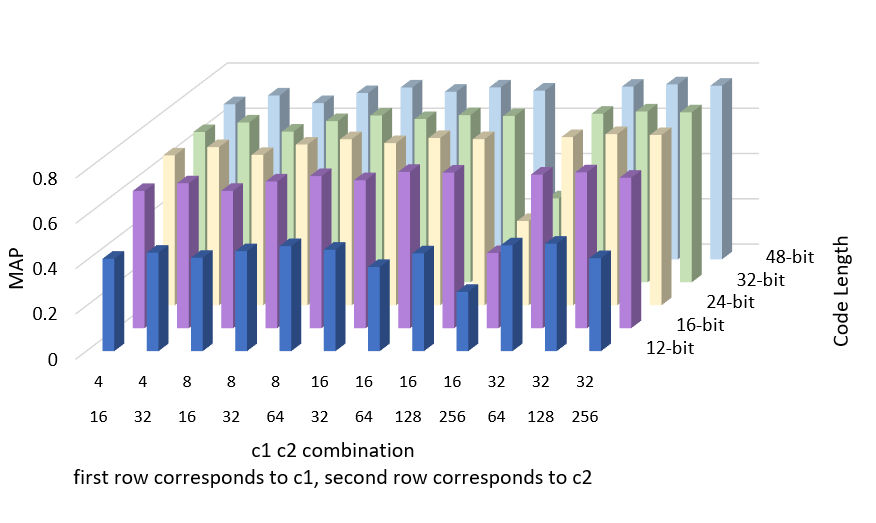}
\caption{Retrieval MAP on CIFAR-10 with different sizes of transition matrixes ( $ c_{1}$ and $ c_{2}$ ) and different length of hash codes.}
\label{fig_sim}
\end{figure}
\par As for different choices of ${{c}_{1}}$ and ${{c}_{2}}$, Figure 7 shows that the performance of the proposed method changes with the change of ${{c}_{1}}$ or ${{c}_{2}}$. However, the changes are not big except for one case with ${{c}_{1}}=16$ and ${{c}_{2}}=256$, whose performance is very bad. This exceptional bad result may result from the too large difference between ${{c}_{1}}$ and ${{c}_{2}}$, where ${{c}_{2}}$ is 8 times as ${{c}_{1}}$. Another observation is that the performance is still satisfactory when ${{c}_{1}}$ and ${{c}_{2}}$ are relatively small. This is very important as the smaller ${{c}_{1}}$ and ${{c}_{2}}$ be, the computational complexity of the following part of the algorithm can be reduced more. This also means that the efficiency the algorithm can be improved without big sacrifice of performance by choosing relatively smaller ${{c}_{1}}$ and ${{c}_{2}}$. It is also noticing that in all cases ${{c}_{2}}$ is set as multiples of ${{c}_{1}}$ considering the size of input feature.
\par From the above discussion on parameters, we can see that our BSDH method is comparatively easy to achieve desired performance by slightly adjusting parameters. This is regarded as another merit of the proposed method as this method is not only effective, but also very robust with parameters.  

\section{Conclusion}
In this paper, we propose a bilinear supervised hashing method named BSDH. The BSDH method is designed based on the idea of directly dealing with native matrix-form image features. It utilizes bilinear projections to directly process the image matrix features, which keeps the image matrix’s intrinsic space relationship and integrates the idea of bilinear projection approximation with vectorization binary codes regression in the objective function. The approach of bilinear projection is also helpful for saving computing memory which can better deal with high dimensional image features than commonly used single linear projection approach. The optimal solutions are obtained by using iteratively updating variables referring to the idea of bilinear discriminant analysis and discrete optimization strategy. Experimental results on four image datasets 
with handcraft and CNN features demonstrate the effectiveness and superiority of the BSDH methods. 

\ifCLASSOPTIONcaptionsoff
  \newpage
\fi
\nocite{*}
\bibliographystyle{IEEEtran}
\bibliography{refe}

\end{document}